\documentclass[pmlr,twocolumn,10pt]{jmlr} 
\usepackage{booktabs}
\usepackage{microtype}
\usepackage{graphicx}
\usepackage{float}
\usepackage{placeins}
\usepackage{colortbl}
\usepackage{listings} 
\usepackage{changepage} 
\usepackage{ragged2e} 
\usepackage{subcaption}
\usepackage{multirow} 
\usepackage{array}  
\usepackage{longtable}
\usepackage{xcolor}
\usepackage{hyperref}
\usepackage{siunitx}
\usepackage{tcolorbox}
\theorembodyfont{\upshape}
\theoremheaderfont{\scshape}
\theorempostheader{:}
\theoremsep{\newline}

\usepackage{soul} 
\usepackage{xcolor} 

\jmlrvolume{287}
\jmlryear{2025}
\jmlrsubmitted{} 
\jmlrpublished{} 
\jmlrworkshop{Conference on Health, Inference, and Learning (CHIL) 2025}

\title[CaReAQA]{CaReAQA: A Cardiac and Respiratory Audio Question Answering Model for Open-Ended Diagnostic Reasoning}

\author{%
\Name{Tsai-Ning Wang}\Email{t.n.wang@tue.nl}\\
\addr Eindhoven University of Technology, The Netherlands
\AND
\Name{Lin-Lin Chen}\Email{L.Chen@tue.nl}\\
\addr Eindhoven University of Technology, The Netherlands
\AND
\Name{Neil Zeghidour}\Email{neil@kyutai.org}\\
\addr Kyutai, France
\AND
\Name{Aaqib Saeed}\Email{a.saeed@tue.nl}\\
\addr Eindhoven University of Technology, The Netherlands \\
Eindhoven Artificial Intelligence Systems Institute, The Netherlands
}

\begin{document}

\maketitle

\begin{abstract}
Medical audio signals, such as heart and lung sounds, play a crucial role in clinical diagnosis. However, analyzing these signals remains challenging: traditional methods rely on handcrafted features or supervised deep learning models that demand extensive labeled datasets, limiting their scalability and applicability. To address these issues, we propose CaReAQA\footnotemark, an audio-language model that integrates a foundation audio model with the reasoning capabilities of large language models, enabling clinically relevant, open-ended diagnostic responses. Alongside CaReAQA, we introduce CaReSound, a benchmark dataset of annotated medical audio recordings enriched with metadata and paired question-answer examples, intended to drive progress in diagnostic reasoning research. Evaluation results show that CaReAQA achieves 86.2\% accuracy on open-ended diagnostic reasoning tasks, outperforming baseline models. It also generalizes well to closed-ended classification tasks, achieving an average accuracy of 56.9\% on unseen datasets. Our findings show how audio-language integration and reasoning advances medical diagnostics, enabling efficient AI systems for clinical decision support.

\end{abstract}
\footnotetext{Dataset and pretrained models are available at \url{https://huggingface.co/datasets/tsnngw/CaReSound} and CaReAQA model at \url{https://huggingface.co/tsnngw/CaReAQA}.}

\begin{figure}[t]
    \centering    
    \includegraphics[width=\columnwidth]{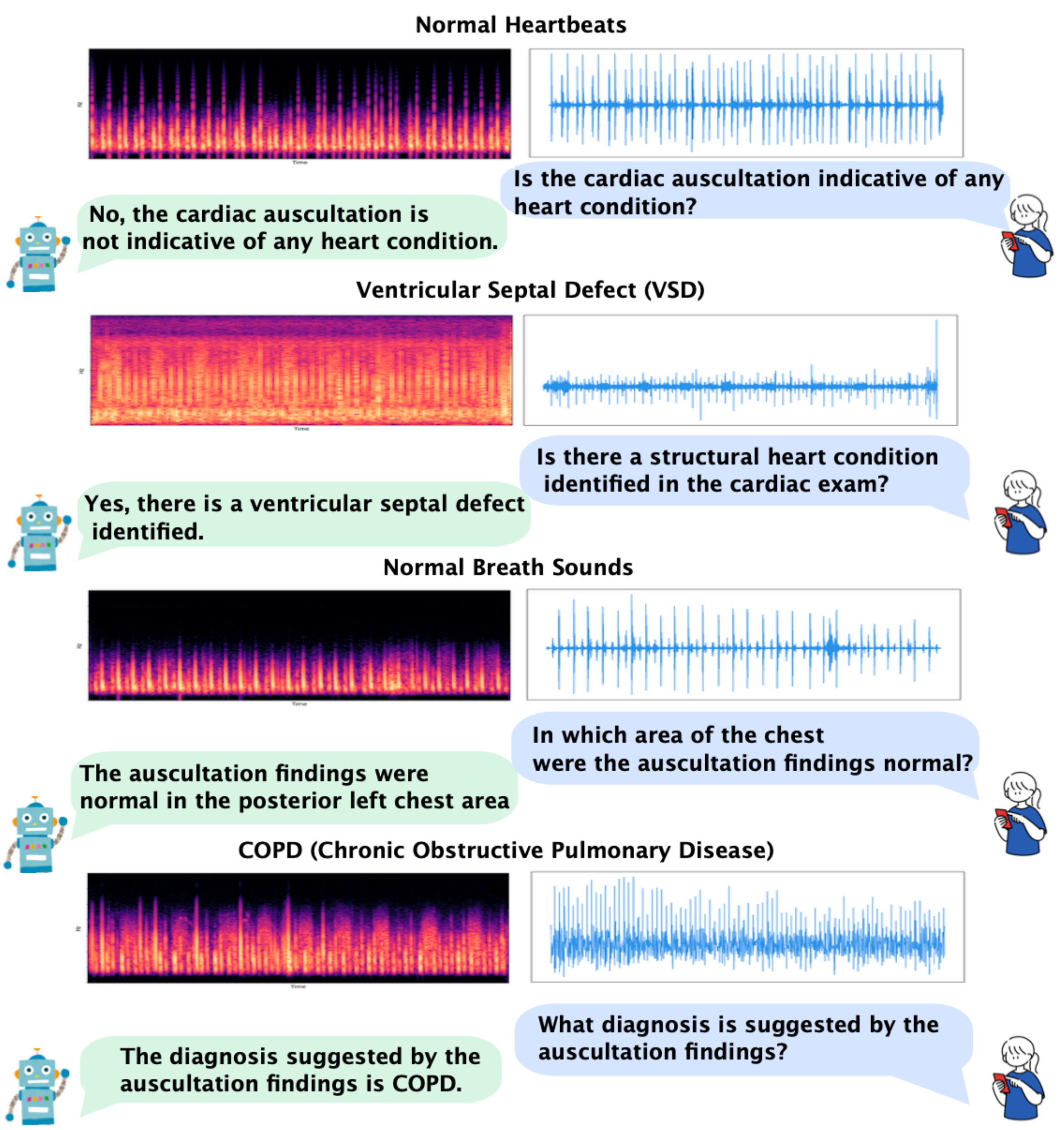} 
    \caption{\small{Spectrograms(left) and waveforms(right) paired with diagnostic question-and-answer outputs, demonstrating the model's ability to analyze and answer questions about normal and abnormal heart and respiratory sounds.}}
    \label{fig:qa}
\end{figure}

\begin{figure*}[t]
    \centering    
    \includegraphics[width=\textwidth]{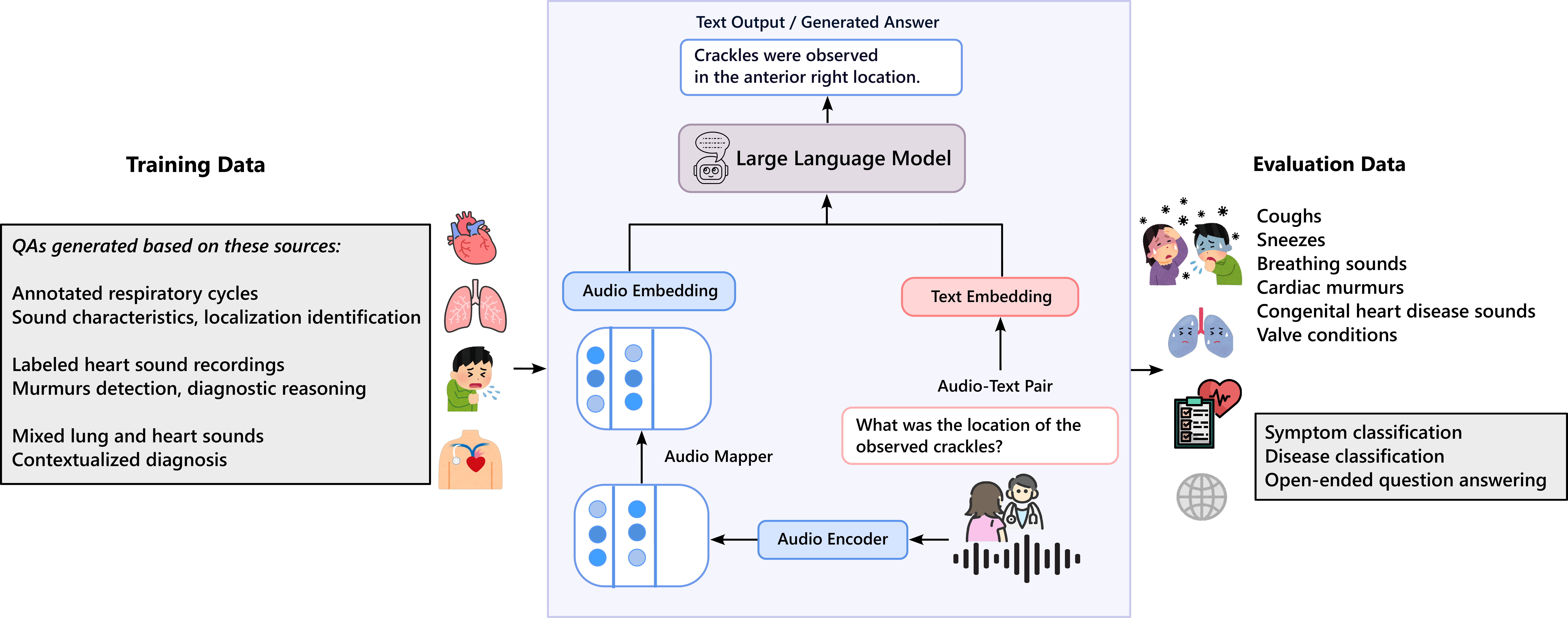} 
    \caption{\small{Overview of CaReAQA. Our multimodal framework combining audio and text data for clinical sound analysis, enabling symptom classification, disease diagnosis, and diagnostic reasoning with a large language model.}}
    \label{fig:arch}
\end{figure*}

\section{Introduction}
\label{sec:intro}

Medical audio data, including cardiac and respiratory sounds, is a rich source of physiological information and plays a crucial role in diagnosing a wide range of health conditions. For example, heart murmurs can indicate structural cardiac abnormalities, while respiratory sounds, such as wheezes or crackles, are often early indicators of conditions like asthma, chronic obstructive pulmonary disease (COPD), or pulmonary fibrosis. Analyzing these sounds more effectively could significantly improve diagnostic accuracy and deepen our understanding of underlying health issues.

Traditional medical audio analysis uses labor-intensive, less adaptable handcrafted features and signal processing techniques. Supervised deep learning models have demonstrated strong performance but require large amounts of annotated data, which are costly and time-consuming to collect. Meanwhile, general-purpose audio-language models, though powerful, lack the domain-specific knowledge needed to capture the unique properties of medical audio. As a result, these models often fall short in identifying critical patterns essential for distinguishing pathological from non-pathological sounds and mainly are limited to close-ended classification tasks.

Despite advancements in multimodal AI, medical audio remains an underexplored domain compared to visual and textual data. Current models are often optimized for broad, non-specialized tasks and insufficiently tailored to the demands of healthcare diagnostics. Furthermore, while versatile models trained on diverse datasets have shown improvements across a wide range of tasks, their lack of specialization limits their utility in clinical applications.

To tackle these challenges, we introduce CaReAQA, a Cardiac and Respiratory Audio Question Answering model tailored for open-ended question answering in medical diagnostics. By integrating a self-supervised audio foundation model with the advanced reasoning capabilities of large language models (LLMs), CaReAQA generates clinically relevant and context-aware diagnostic responses that adapt to the complexity of real-world medical scenarios. Unlike traditional models limited to predefined tasks or fixed outputs, CaReAQA provides the flexibility and depth required to address diverse and nuanced diagnostic challenges effectively.

Our contributions are as follows:
\begin{itemize}
    \item We introduce a novel audio-language model designed specifically for medical diagnostic question-answering task, integrating self-supervised audio encoder with the reasoning capabilities of large language models to produce open-ended diagnostic answers.
    \item We introduce CaReSound, a new benchmark dataset comprising of diverse public medical audio recordings, such as respiratory and cardiac sounds, annotated with detailed metadata and question-answer pairs. 
    \item We establish a robust evaluation framework to assess the model's performance across a variety of tasks, including open-ended question answering and closed-ended classification tasks.
\end{itemize}

Our experimental evaluation demonstrates that CaReAQA significantly outperforms strong baselines on open-ended diagnostic reasoning tasks across diverse datasets. By accurately interpreting complex patterns in medical audio and generalizing effectively to unseen data, our model addresses the unique challenges of medical diagnostics that general-purpose audio models fail to meet. Our work not only lays the groundwork for advanced diagnostic tools but also highlights its potential to transform health monitoring by enabling accurate, efficient, and accessible diagnostic support, ultimately improving patient outcomes and aiding clinicians in decision-making. We emphasize that CaReAQA is not intended for direct clinical deployment but a step toward advancing auscultation-based diagnostic reasoning.

\section{Related Work}

\label{related_works}
\subsection{General-Purpose Audio-Language Models}

Recent advances in audio-language models have significantly enhanced the integration of auditory perception and reasoning, enabling LLMs to handle diverse audio-related tasks. For instance, the LTU (Listen, Think, Understand) model \citep{gong2023listen} excels in both closed and open-ended question-answering tasks.

Instruction-based frameworks like Pengi \citep{deshmukh2023pengi} improve task versatility by framing diverse tasks as instruction-response problems, while models like GAMA \citep{ghosh2024gama} and AudioGPT \citep{huang2024audiogpt} leverage pre-trained audio models for reasoning over complex auditory inputs. AudioGPT, in particular, excels in multi-turn dialogues for general audio tasks but relies on predefined interfaces like ChatGPT and external systems (e.g., ASR or task-specific models) for audio processing.

In contrast, our model natively processes audio and text inputs within a unified framework, eliminating reliance on external systems or pipelines. By focusing on medical audio data, it addresses the unique challenges of health diagnostics and generates tailored open-ended responses, bridging a critical gap left by general-purpose models.

\subsection{Domain-Specific Audio and Multimodal Applications}
Domain-specific applications of audio-language models remain relatively under-explored. Efforts like RespLLM \citep{zhang2024respllm} have begun addressing this gap by leveraging pretrained large language models and cross-modal attention mechanisms to fuse audio and text representations for classification tasks. However, its scope remains focused on respiratory sounds, leaving broader health diagnostics unaddressed. Likewise, RespLLM is limited to binary classification tasks, providing only 1 (``yes'') or 0 (``no'') outputs through a basic linear layer without support for question answering. In contrast, our work is built to handle open-ended question answering. This broader focus makes our approach more adaptable to complex challenges in audio-based health monitoring, going beyond the narrow scope of binary classification.

OPERA \citep{NEURIPS2024_2f803abd} introduces a self-supervised framework for respiratory audio, leveraging large-scale unlabeled datasets to learn robust acoustic representations. While effective for tasks like health condition inference and lung function estimation, it relies on training or fine-tuning a classifier on top of the encoder and remains limited to respiratory audio. Additionally, it lacks the ability to perform audio-language reasoning or generate open-ended diagnostic responses. In contrast, our approach overcomes these limitations and extends to cardiac sounds, enabling a broader and more integrated analysis of acoustic biomarkers.

Medical visual question answering (VQA) systems have demonstrated significant potential in automating clinical reasoning by answering questions based on medical images \citep{van2023open}. However, while these systems excel at leveraging visual data for open-ended reasoning, they overlook the diagnostic value of auditory inputs, such as respiratory and cardiac sounds.

General multimodal large language models, such as Gemini \citep{geminiteam2024geminifamilyhighlycapable}, QWen-Audio \citep{chu2023qwen} and GPT-4o \citep{openai2024gpt4technicalreport}, extend reasoning capabilities to include vision, language, and auditory modalities. However, challenges remain in addressing domain-specific auditory tasks that require specialized expertise. Our approach emphasizes the use of medical audio inputs to provide open-ended answers, thereby supporting diagnostic decision-making in healthcare scenarios.

\section{Methodology}
\subsection{Problem Statement}
In this work, we aim to develop an audio-language model capable of generating open-ended diagnostic answers using medical audio data and natural language inputs. The primary challenge is to effectively combine the audio understanding capabilities with the reasoning power of large language models (LLMs) to produce meaningful answers to the posed questions. Given an audio input \( X_a \) and a related question \( X_q \) in natural language, our objective is to generate an answer sequence \( A = \{A_0, A_1, \dots, A_N\} \) that accurately conveys diagnostic insights. Formally, the task can be represented as finding the optimal model parameters \( \theta^* \) by maximizing the conditional likelihood, \( A \) as follows:
\[
\theta^* = \arg \max_{\theta} \sum_{i=1}^{N} \log p_{\theta}(A_i \mid X_q, X_a, A_{i-1}),
\]

\noindent where \( p_{\theta} \) represents the probability of each token in the answer sequence, conditioned on the given question, audio input, and previously generated tokens.

\subsection{CaReSound Benchmark Dataset}
\begin{table*}[!htbp]
\centering
\footnotesize
\captionsetup{skip=5pt}
\caption{\small{Comprehensive statistics of the in-domain datasets used for CaReSound benchmark. The table presents metrics such as sample counts, average durations, question-answer pairs, and the defining characteristics of sound types (e.g., lung and heart sounds) along with their associated medical conditions.}}
\label{training-datasets-table}
\begin{tabular}{lcccccc}
\toprule
\textbf{} & \textbf{ICBHI} & \textbf{KAUH} & \textbf{CirCor} & \textbf{ZCH} & \textbf{SPR} & \textbf{All} \\
\midrule
Number of Samples          & 6899   & 337   & 5282   & 1259   & 2496   & 16273    \\
Duration (s)               & 22.2   & 17.49 & 22.87  & 20.06  & 11.15  & 18.75    \\
QA Pairs                   & 20729  & 1010  & 5497   & 2527   & 5029   & 34792    \\
Number of Questions        & 20728  & 1009  & 3284   & 1477   & 2504   & 29002    \\
Mean Length of Questions   & 10.03  & 7.61  & 9.13   & 9.76   & 9.28   & 9.56     \\
Mean Length of Answers     & 6.22   & 2.07  & 9.31   & 10.37  & 7.70   & 7.53     \\
Number of Unique Answers   & 5614   & 95    & 2113   & 1476   & 2060   & 11358    \\
\midrule
Sound Type                 & Lung   & Lung  & Heart  & Heart  & Lung   & Mixed    \\
Description                & COPD, etc. & Asthma, etc. & Murmurs & CHD & Pediatric sounds & Mixed \\
\bottomrule
\end{tabular}
\end{table*}

\begin{table}[!htbp]
\centering
\captionsetup{skip=5pt}
\caption{\small{Statistics of out-of-domain evaluation datasets used in the CaReSound benchmark evaluation.}}
\label{evaluation-datasets-table}
\resizebox{\columnwidth}{!}{%
\begin{tabular}{lccccc}
\toprule
\textbf{} & \textbf{BMD} & \textbf{FluSense} & \textbf{Coswara} & \textbf{TR} & \textbf{All} \\
\midrule
Number of Samples          & 872    & 758    & 2746   & 504    & 4880    \\
Classes                    & 5      & 9      & 4      & 2      & -       \\
Mean Duration (s)          & 19.99  & 20.89  & 15.92  & 21.75  & 18.92   \\ 
Sound Type                 & Heart  & Respiratory & Respiratory & Lung   & Mixed   \\
Description                & CVD    & Sneeze, etc. & Cough, etc. & COPD  & All datasets \\
\bottomrule
\end{tabular}
}
\end{table}

One of our key contributions is the creation of a novel benchmark dataset, CaReSound, built from a diverse collection of open-source medical audio recordings for question answering task. The included datasets are as follows:

\begin{itemize}
    \item \textbf{ICBHI} \citep{DVN/HT6PKI_2023}: The Respiratory Sound database, created for the ICBHI 2017 scientific challenge, includes recordings with 6,898 annotated respiratory cycles from 126 subjects. These cycles feature crackles, wheezes, combinations of both, or no adventitious sounds.

    \item \textbf{KAUH} \citep{10404405}: Respiratory sounds from 112 subjects, with recordings containing at least one respiratory cycle. Each recording is annotated with detailed metadata, including diagnosis, sound type, chest zone, and subject demographics.

    \item \textbf{CirCor} \citep{oliveira2021circor}: Pediatric heart sound recordings from 1,568 subjects, with detailed annotations of murmurs and heart sound segmentation.

    \item \textbf{SPRSound} \citep{zhang2022sprsound}: A pediatric respiratory sound database of annotated respiratory recordings from 292 participants. It includes detailed annotations at both record and event levels for classifications of respiratory sounds.

    \item \textbf{ZCHSound} \citep{jia2024zchsound}: A pediatric heart sound database containing recordings from 1,259 participants, including 566 congenital heart disease cases.
\end{itemize}

These recordings, primarily consisting of respiratory and cardiac sounds, are accompanied by detailed metadata that provides descriptions of the audio content, such as sound types, user demographics, and diagnostic details. 

\looseness=-1
To generate diverse and contextually relevant question-and-answer (QA) pairs for training and evaluation, we adopt a methodology leveraging large language models (LLMs). Generating open-ended QA pairs poses significant challenges, as manually creating hundreds of thousands of such pairs is both time-consuming and impractical. To address this, we leverage off-the-shelf LLMs to automatically generate QA pairs, offering an efficient and scalable solution for producing high-quality data suitable for fine-tuning, inspired by recent efforts in creating diverse, domain-specific datasets for fine-tuning LLMs in specialized fields.

We employ GPT-4o \citep{openai2024gpt4technicalreport} to generate question-answer (QA) pairs based on metadata and annotations extracted from the datasets. We exclusively utilize textual meta-information.

Specifically, the key metadata varies per dataset and commonly includes subject demographics, recording locations, and diagnostic labels. We incorporate it into carefully designed prompts (see Appendix  \ref{apd:appendix-prompt} for the complete prompt example), enabling GPT-4o to generate QA pairs that reflect the diagnostic and clinical relevance of the audio recordings.

This method highlights the adaptability of large language models to audio-related tasks by leveraging textual representations instead of raw signal data. For each audio recording, we generate multiple QA pairs to capture diverse clinical and diagnostic contexts. Table~\ref{training-datasets-table} provides a detailed summary of the dataset, including the number of samples, total duration, and the number of QA pairs generated. Furthermore, representative examples of QA pairs used during training can be found in Appendix \ref{apd:training_data} , Table~\ref{qa-sample-table}.

\subsection{Evaluation Data}
We evaluate the model on two key tasks: open-ended question-answering and closed-ended classification. These tasks are designed with distinct settings to rigorously assess the model's generalization capabilities as well as its performance on both in-domain and out-of-domain data.

\subsubsection{Open-Ended Question Answering}

The open-ended question answering task is evaluated on the designated test splits of the datasets, ensuring no overlap with the training data. While the evaluation sets originate from the same datasets, the multi-dataset training framework introduces additional complexity. For example, questions from different datasets often use varied terminology tailored to specific medical domains. One dataset might emphasize respiratory sounds with terms like ``crackles'' or ``wheezes,'' while another focuses on cardiac murmurs, requiring familiarity with phrases like ``holosystolic murmur'' or grading descriptions such as ``III/VI.'' This diversity in language requires the model to adapt dynamically to each context. Additionally, some datasets contain concise answers such as ``No crackles detected'', while others include elaborate diagnostic explanations like ``A holosystolic, high-pitched murmur detected over the mitral area''. Handling these variations in response length and detail adds another layer of complexity for model generalization. Unlike models tailored to individual datasets or tasks, our approach avoids dataset-specific fine-tuning or task-specific adaptations, providing a robust assessment of the model’s ability to generalize across diverse input-output mappings within its domain.

\looseness=-1
The evaluation assesses the model's ability to generate accurate and clinically relevant diagnostic answers. We use BERTScore \citep{zhang2019bertscore}, a metric that quantifies semantic similarity between generated responses and reference answers. Implementation of BERTScore leverages Hugging Face's ``evaluate'' library with the pre-trained ``bert-base-uncased model''. To measure lexical similarity while accounting for stemming, synonyms, and word order, we include METEOR \citep{banerjee2005meteor} as an additional metric. In some cases, an incorrect diagnosis may have a high similarity with the correct one, due to subtle lexical differences, negations, etc. To circumvent this issue, we compute an Accuracy score by comparing the model's predictions against the ground truth using GPT-4o \citep{openai2024gpt4technicalreport}. We prompt the GPT-4o model via API to compare whether the prediction and ground truth are the same, providing a binary judgment ``Yes'' or ``No'' for each pair, indicating whether the generated answer and the reference are semantically identical. Details of the evaluation prompt used with GPT-4o can be found in  Appendix~\ref{apd:accuracy-eval}. We aggregate these binary judgments to compute the model's overall accuracy.

\subsubsection{Generalization to Unseen Data}

The model's generalization capabilities are further rigorously evaluated on multiple unseen datasets of medical audio recordings annotated with diagnostic labels. This experimental design eliminates any overlap with the training domain data and ensures a comprehensive assessment of the model’s ability to handle entirely novel out-of-domain scenarios.

The evaluation spans four diverse datasets:

\textbf{TR} \citep{altan2020respiratory}, which contains lung sound recordings from 75 subjects, along with pulmonary function tests, classified into two categories: COPD or not (2 classes).

\textbf{BMD-HS} \citep{ali2024buet}, featuring heart sound recordings from 59 patients, annotated into six categories, including valvular heart diseases such as aortic stenosis and mitral regurgitation, as well as healthy (normal) cases (6 classes).

\textbf{Coswara} \citep{bhattacharya2023coswararespiratorysoundssymptoms}, comprising respiratory sounds from 2,635 individuals, classified based on demographic and behavioral attributes: smoker or non-smoker, and male or female (4 classes).

\textbf{FluSense} \citep{al2020flusense}, derived from AudioSet \citep{gemmeke2017audio}, containing audio labeled for events like coughs, sneezes, sniffles, and throat-clearing, categorized into 9 distinct classes.

Key characteristics of these datasets, including sample counts, class distribution, mean duration, sound types, and description, are summarized in Table~\ref{evaluation-datasets-table}. The model performance is quantified using accuracy, computed as the proportion of correct predictions. 

\subsection{Model Architecture}
Our model follows an encoder-decoder framework that integrates medical audio and textual questions using a large language model (LLM) to generate diagnostic answers. It comprises three main components: an audio feature extractor, a text embedder, and the decoder-only LLM for multimodal input processing and output generation.

The audio feature extractor processes the input medical audio \( X_a \) by first converting it into a log-mel spectrogram to capture its time-frequency characteristics. An audio encoder extracts features from the spectrogram, resulting in audio embeddings represented as:
\[
Z_a \in \mathbb{R}^{L_a \times A},
\]

\noindent where \( L_a \) is the sequence length, and \( A \) is the feature dimension of the audio embeddings. These embeddings capture the essential characteristics of the medical audio relevant for diagnostic reasoning. To effectively integrate the information from both modalities, the audio embeddings are designed to be aligned with the text embeddings during multimodal fusion. 

Simultaneously, the text embedder processes both the question \( X_q \) and the answer \( A \) during training. The question is tokenized into a sequence \( \{q_1, q_2, \dots, q_{L_q}\} \), and the answer into \( \{a_1, a_2, \dots, a_{L_a}\} \). These token sequences are embedded using the LLM's embedding layer, yielding:
\[
Z_q \in \mathbb{R}^{L_q \times S}, \quad Z_A \in \mathbb{R}^{L_a \times S},
\]
\noindent where \( S \) is the shared embedding dimension. The embeddings \( Z_q \), \( Z_A \), and \( Z_a \) are combined later in the pipeline to form a unified multimodal representation, allowing the model to learn relationships between the question, audio features, and the answer.

To achieve multimodal fusion, the audio embeddings \( Z_a \) and text embeddings \( Z_q \) are concatenated to form a unified representation:
\[
Z = [Z_q; Z_a] \in \mathbb{R}^{L \times S}, \quad L = L_q + L_a.
\]
Positional embeddings \( P \in \mathbb{R}^{L \times S} \) are added to \( Z \) to preserve the temporal and semantic order of the tokens:
\[
Z_{\text{final}} = Z + P.
\]

The combined representation \( Z_{\text{final}} \) is processed by the LLM's Transformer layers, which employ self-attention mechanisms to integrate the multimodal information. The output is then passed through a softmax layer to generate the diagnostic answer probabilities:
\[
P(A | X_q, X_a) = \text{Softmax}(\text{Transformer}(Z_{\text{final}})).
\]

During training, the model maximizes the conditional likelihood of the answer sequence given the input audio \( X_a \) and the question \( X_q \). The objective function is defined as:
\[
\mathcal{L} = -\sum_{t=1}^{T} \log P(A_t | X_q, X_a, A_{<t}),
\]
\noindent where \( A_{<t} \) represents the tokens generated previously up to step \( t \).

\subsection{Training and Evaluation Process}

The model is trained and evaluated on the CaReSound benchmark dataset we curated, which consists of a training set (\( D_{\text{train}} \)) and a test set (\( D_{\text{test}} \)). The dataset contains 12,673 samples, which we split into 80\% for training and 20\% for testing, resulting in 10,138 training samples and 2,535 test samples. There is no overlap between these splits both in terms of audio recordings and patients. During training, the model learns to generate diagnostic answers by predicting a sequence of tokens that match the ground-truth answer. This process is guided by a cross-entropy loss function, where the likelihood of each token in the answer is maximized given the audio input, the question, and the tokens generated so far.

The loss function is computed across all samples in the training set and over all tokens in the target sequence. For a given input comprising an audio signal \( x_a \), a natural language question \( x_q \), and the corresponding target sequence \( a = \{a_1, a_2, \dots, a_N\} \), the loss for each token \( a_i \) is defined as:
\[
\log p_\theta(a_i \mid x_a, x_q, a_{<i}),
\]
where \( a_{<i} \) denotes the tokens generated prior to \( a_i \), and \( \theta \) represents the trainable parameters of the model. 

To evaluate the model’s robustness, testing is conducted on the designated evaluation datasets  (\( D_{\text{test}} \)) The evaluation process is designed to assess the model's performance across different task types. For classification tasks, the model predicts outcomes for diagnostic classes, aligning responses with categories. Accuracy is used as the primary evaluation metric, measuring the proportion of correct predictions. For open-ended question answering, the model generates free-form responses that are assessed using semantic and lexical similarity metrics, such as BERTScore, METEOR, and Accuracy.

\noindent 
\section{Experiments}
\begin{table*}[t]
\centering
\small
\caption{Performance comparison of BertScore (BertS.), METEOR (MET.), and Accuracy (Acc.) in an open-ended question answering task. The ``All*'' column aggregates results from all datasets.}
\label{open-ended}
\resizebox{\textwidth}{!}{%
\begin{tabular}{l
                S[table-format=2.1] S[table-format=2.1] S[table-format=2.2]
                S[table-format=2.1] S[table-format=2.1] S[table-format=2.2]
                S[table-format=2.1] S[table-format=2.1] S[table-format=2.2]
                S[table-format=2.1] S[table-format=2.1] S[table-format=2.2]
                S[table-format=2.1] S[table-format=2.1] S[table-format=2.2]
                S[table-format=2.1] S[table-format=2.1] S[table-format=2.2]}
\toprule
\textbf{Method} & \multicolumn{3}{c}{\textbf{ICBHI}} & \multicolumn{3}{c}{\textbf{CIRCOR}} & \multicolumn{3}{c}{\textbf{KAUH}} & 
\multicolumn{3}{c}{\textbf{SPRSound}} & \multicolumn{3}{c}{\textbf{ZCHSound}} & \multicolumn{3}{c}{\textbf{All*}} \\
\cmidrule(lr){2-4} \cmidrule(lr){5-7} \cmidrule(lr){8-10} \cmidrule(lr){11-13} \cmidrule(lr){14-16} \cmidrule(lr){17-19}
& {BertS} & {MET.} & {Acc.} & {BertS} & {MET.} & {Acc.} & {BertS} & {MET.} & {Acc.} & 
{BertS} & {MET.} & {Acc.} & {BertS} & {MET.} & {Acc.} & {BertS} & {MET.} & {Acc.} \\
\midrule
Majority Answer & 50.9 & 6.2 & 12.0 & 75.3 & 24.0 & 1.8 & 62.5 & 10.2 & 20.5 & 77.8 & 33.3 & 5.4 & 73.4 & 24.4 & 5.5 & 67.9 & 19.6 & 9.0\\
LLM w/o audio & 44.4 & 23.6 & 7.7 & 49.1 & 29.2 & 10.0 & 34.2 & 0.3 & 4.9 & 48.9 & 21.5 & 10.1 & 54.0 & 39.5 & 10.1 & 49.4 & 24.6 & 7.8 \\
Cascaded & 41.5 & 17.6 & 9.1 & 47.6 & 20.9 & 23.7 & 31.8 & 0.2 & 39.1  & 44.3 & 17.0 & 15.8 & 51.3 & 26.9 & 37.8 & 43.6 & 17.8  & 17.0 \\
LTU \citep{gong2023listen} & 45.6 & 14.2 & 24.0 & 49.4 & 20.2 & 23.7 & 34.4 & 0.1 & 5.4 & 47.6 & 18.3 & 44.3 & 49.8 & 21.5 & 38.0 & 51.0 & 14.7 & 24.5 \\
Pengi \citep{deshmukh2023pengi} & 37.0 & 1.0 & 1.0 & 33.6 & 1.4 & 1.0 & 40.0 & 0.2 & 0.1 & 35.0 & 0.9 & 1.1 & 34.3 & 2.5 & 0.7 & 37.4 & 1.1 & 0.9 \\
Qwen2-Audio \citep{chu2023qwen} & 47.2 & 23.4 & 17.1 & 48.2 & 29.8 & 18.8 & 34.6 & 0.3 & 7.9 & 48.3 & 25.4 & 32.7 & 49.5 & 26.8 & 19.0 & 53.5 & 30.2 & 17.0 \\
Gama \citep{ghosh2024gama}& 59.4 & 27.5 & 17.9 & 62.2 & 30.6 & 33.6 & 39.1 & 0.3 & 5.9 & 60.3 & 29.5 & 16.6 & 64.6 & 34.2 & 26.3 & 56.8 & 28.4 & 18.2 \\
\midrule
\textbf{CaReAQA (Ours)} & \textbf{82.2} & \textbf{67.0} & \textbf{72.5} & \textbf{87.2} & \textbf{79.0} & \textbf{49.1} & \textbf{74.7} & \textbf{50.2} & \textbf{23.9} & \textbf{86.6} & \textbf{78.5} & \textbf{76.2} & \textbf{88.5} & \textbf{91.3} & \textbf{82.1} & \textbf{86.2} & \textbf{77.5} & \textbf{70.6} \\
\bottomrule
\end{tabular}%
}
\end{table*}

\subsection{Training Recipe}
Our model employs LLaMA-3.2-3B \citep{llama2023} as the default large language model, fine-tuned with LoRA using a rank of 8 as the default configuration unless otherwise stated. For audio feature extraction, we fine-tuned the OPERA encoder \citep{NEURIPS2024_2f803abd} to enable audio model learn representations of both lung and heart sounds.

To bridge the audio encoder and the language model, we introduce a transformer-based mapping network. This mapper leverages a multi-head self-attention mechanism, a feed-forward network, pre-norm layer normalization, and positional encodings to transform temporal audio representations into a format compatible with the LLM's input dimensions.

We train the model using the AdamW optimizer with a batch size of 64 and a linear learning rate schedule. The learning rate is set to 
\(2 \times 10^{-5}\), with 400 warmup steps to ensure stable and efficient convergence. To address memory constraints, we employ gradient accumulation, enabling parameter updates after processing multiple batches. The model is trained over 50 epochs, with each training run taking approximately one day on an NVIDIA A100 GPU. 

During training, audio inputs are segmented into 5-second clips on the fly. For each batch, a random 5-second segment is extracted from each audio clip to account for variability in recording durations, ensuring consistent input lengths across batches. To enhance dataset diversity, we apply data augmentation using the \texttt{AugLy} library \citep{DBLP:journals/corr/abs-2201-06494}. Augmentations are randomly and selectively applied to individual audio clips within each batch, introducing variability while preserving diagnostic features. Specifically, we apply one of four transformations with equal probability: a 5dB volume increase, amplitude normalization, a low-pass filter (cutoff at 300Hz), or a high-pass filter (cutoff at 3000Hz). 

\subsection{Open-Ended Audio Task Experiments}
The open-ended diagnostic reasoning task evaluates the ability of models to generate accurate and contextually relevant answers across diverse datasets part of CaReSound benchmark. As summarized in Table~\ref{open-ended}, CaReAQA consistently outperforms all baseline methods, demonstrating robust generalization across different domains and datasets. This section details the evaluation framework and the performance of the baseline models considered.

While CaReAQA shows strong diagnostic reasoning, challenges remain with overlapping auscultation patterns. See Appendix~\ref{apd:failure_cases} for failure case analysis.

\textbf{Majority Answer Baseline.}
We begin with evaluating the ``Majority Answer'' approach, where each question is answered using the most frequent response from the training set of the corresponding dataset. Performance varies significantly across datasets due to differences in answer structure and diversity. In KAUH, the majority answer is a single word (``Normal''), resulting in lower performance on metrics that prioritize detailed and nuanced responses. Conversely, datasets like CIRCOR and ZCHSound, where the majority answers are longer and more descriptive (e.g., ``No, no murmurs were detected during the cardiac assessment''), achieve higher scores due to greater alignment with the ground truth. Nevertheless, accuracy remains limited in datasets with high variability in responses, as a single majority answer fails to capture the full diversity of the data.

\textbf{Cascaded Baseline.} 
Next, we evaluate a cascaded baseline that separates audio feature extraction and reasoning into two stages. First, a linear classification head is added to the audio encoder and fine-tuned on labeled datasets (e.g., ZCH heart sounds labeled with Atrial Septal Defect, SPR respiratory sounds labeled with Rhonchi and Wheeze) using cross-entropy loss. The model is trained with AdamW for 50 epochs at a learning rate of \(2 \times 10^{-5}\) and a batch size of 64. 

\looseness=-1
In the second stage, predicted labels are used as prompts for LLaMA-3.2-3B without further fine-tuning. For example, in adventitious lung sound classification, the prompt might be: ``What kind of adventitious lung sounds are noted in this examination? The diagnosis is [predicted label],'' where \texttt{[predicted label]} (e.g., ``Bronchiectasis'') is replaced dynamically.

Although this modular setup provides flexibility by decoupling audio classification from reasoning, it lacks the benefits of end-to-end optimization. As a result, the cascaded baseline performs significantly worse than our proposed CaReAQA model. For instance, it achieves an accuracy of only 37.8\% on the ZCHSound dataset, compared to 82.1\% achieved by CaReAQA. 

\textbf{Comparison with Audio-Language Models.}
We evaluated our approach against several state-of-the-art audio language models, including LTU, Pengi, Qwen2-Audio, GAMA, and LLaMA-3.2-3B (used without audio input). LLaMA-3.2-3B serves as the primary baseline for non-audio evaluation due to its optimal balance of efficiency and accuracy, as established in our preliminary evaluations. Detailed performance comparisons for other large language models (including instruct-tuned versions) are provided in Appendix~\ref{apd:model_performance} and Appendix~\ref{apd:instruction-tuned}. 

While LTU and Pengi are designed for general-purpose question answering, they fail to perform effectively on medical datasets due to limited exposure to clinical terminology and auscultation sounds. For instance, Pengi produces irrelevant responses, such as ``feet maintaining mic'', when prompted with questions about specific auscultatory findings. Similarly, LTU generates overly generic answers like ``No, only a heartbeat can be heard'', which lack clinical utility and precision.

Qwen2-Audio and GAMA outperform LTU and Pengi but exhibit significant limitations in generating accurate and detailed diagnostic outputs. For instance, on the KAUH dataset, GAMA misidentifies respiratory sounds, providing vague statements such as ``Yes, faint crackling sounds are presen'', even when the ground truth indicates no adventitious sounds.
In contrast, our model demonstrates consistent superiority across multiple datasets, achieving an average BERTScore of 86.2 and METEOR score of 77.5, compared to GAMA's 56.8 and 28.4, respectively.

\textbf{Evaluation on the Combined Dataset.}

We evaluate overall performance using the combined dataset (All) rather than relying on macro averaging. This approach directly measures the model's ability to handle a diverse range of examples spanning all datasets, offering a more accurate representation of real-world scenarios where data are not neatly partitioned by task or domain. In contrast, macro-averaging computes performance metrics separately for each dataset and averages them equally, regardless of the dataset sizes. While this ensures that smaller datasets are not overshadowed by larger ones, it can amplify the influence of noise or outliers in less-representative datasets and obscure trends in larger datasets. By evaluating on the combined dataset, we provide a comprehensive and practical evaluation of the model's open-ended QA capabilities.

\begin{table}[t]
\centering
\caption{Performance comparison of accuracy (\%) across different datasets for closed-ended audio tasks.}
\label{classification-table}
\resizebox{\columnwidth}{!}{%
\begin{tabular}{lccccc}
    \toprule
    \textbf{Method} & \textbf{TR} & \textbf{Coswara} & \textbf{BMD} & \textbf{FluSense} & \textbf{Average} \\
    \midrule
    LLM w/o audio   & 30.0 & 25.4 & 20.8 & 25.7 & 25.5  \\
    LTU    \citep{gong2023listen}         & 34.2 & 35.7 & 25.9 & 29.9 & 31.4  \\
    Pengi       \citep{deshmukh2023pengi}    & 49.3 & 26.7 & 54.1 & 28.0 & 39.5  \\
    Qwen2-Audio \citep{chu2023qwen}    & 40.2 & 52.5 & 51.0 & 43.1 & 46.7  \\
    Gama  \citep{ghosh2024gama}          & 33.8 & 30.3 & 29.8 & 27.5 & 30.3 \\
    \midrule
    \textbf{CaReAQA (Ours)}   & \textbf{53.5} & \textbf{54.4} & \textbf{73.3} & \textbf{46.5} & \textbf{56.9} \\
    \bottomrule
\end{tabular}
} 
\end{table}

\subsection{Close-Ended Task on Unseen Data.}

The generation of accurate open-ended diagnostic answers relies on the model’s ability to perform fundamental classification tasks. As an open-ended audio question answering system, CaReAQA exhibits strong generalization to closed-ended tasks. We evaluate CaReAQA on classification tasks using previously unseen datasets. Table~\ref{classification-table} summarizes its performance across multiple datasets, including TR, Coswara, BMD, and FluSense. CaReAQA achieves an average accuracy of 56.9\%, consistently outperforming baseline models on the majority of datasets, highlighting its robust generalization capabilities.

\begin{table}[t]
    \centering
    \caption{Performance comparison (\%) across various training configurations.}
    \label{training-configurations}
    \scriptsize
    \begin{tabular}{lcc}
    \toprule
    \textbf{Training Setting} & \textbf{BertScore} & \textbf{METEOR} \\
    \midrule
    No LoRA & 84.3 & 75.4 \\
    VeRA & 76.4 & 71.1 \\
    Mapper & 81.3 & 66.3 \\ 
    LoRA & \textbf{86.2} & \textbf{77.5} \\ \midrule
    Frozen Audio Encoder & 84.8 & 75.4 \\
    Fine-tuned Audio Encoder & \textbf{86.2} & \textbf{77.5} \\
    \bottomrule
    \end{tabular}
\end{table}

\textbf{Performance Across Datasets.}
CaReAQA achieves the highest accuracy on the BMD dataset (73.3\%), which focuses on diagnosing specific cardiac conditions, such as mitral or aortic valve diseases. The structured nature of this dataset, with questions explicitly targeting diagnoses (e.g., ``Does the patient have Aortic Regurgitation?''), establishes a strong correlation between audio features and diagnostic labels, enabling correct predictions. On the TR dataset, CaReAQA also performs strongly in classifying whether a patient has COPD, demonstrating its ability to analyze complex audio patterns and identify key features indicative of the condition. Likewise, for Coswara and FluSense, which involve detecting respiratory symptoms such as coughing, sneezing, or throat clearing, CaReAQA delivers competitive performance. These datasets emphasize the model’s ability to interpret and classify distinct sound patterns, even when the tasks require discerning subtle variations in audio features.

\begin{table}[t]
    \centering
    \caption{Performance comparison of different PEFT methods. The scores are reported as percentages (\%).}
    \label{peft-methods}
    \scriptsize
    \begin{tabular}{lcc}
    \toprule
    \textbf{PEFT Setting} & \textbf{BertScore} & \textbf{METEOR} \\
    \midrule
    Frozen                &   81.3  &   66.3 \\
    Prefix Tuning         &   69.0  &   50.6 \\
    Prompt Tuning         &   38.1  &   18.9 \\
    P-Tuning              &   54.5  &   37.8 \\
    LoRA                  & \textbf{86.2} & \textbf{77.5} \\
    \bottomrule
    \end{tabular}
\end{table}

\textbf{Comparison with Baselines.}

Baseline models show inconsistent performance across datasets, highlighting their limited ability to generalize to closed-ended tasks across unseen data. For instance, Pengi achieves relatively high accuracy on the BMD dataset (54.1\%) but performs poorly on Coswara (26.7\%). Similarly, Qwen2-Audio demonstrates strong results on Coswara (52.5\%) and FluSense (43.1\%) but struggles with datasets like BMD that demand more specialized reasoning. While LTU and GAMA achieve moderate accuracy on certain datasets, they fall short of matching CaReAQA's overall consistency and robust performance across tasks.

\begin{table}[!htbp]
    \centering
    \caption{Impact of varying LoRA rank \(r\) on model performance, with scores reported as percentages (\%).}
    \label{lora-rank}
    \scriptsize
    \begin{tabular}{lcc}
    \toprule
    \textbf{LoRA Rank} & \textbf{BertScore} & \textbf{METEOR} \\
    \midrule
    \( r=4 \)  & 82.2 & 75.9 \\
    \( r=8 \)  & \textbf{86.2} & \textbf{77.5} \\
    \( r=16 \) & 85.6 & 75.6 \\
    \bottomrule
    \end{tabular}
\end{table}

\subsection{Ablation Studies}

\textbf{Impact of Training Configurations.} 
Table~\ref{training-configurations} highlights the performance differences across various training configurations. The No LoRA setting, where only the encoder and mapper are fine-tuned, significantly underperforms compared to configurations incorporating LoRA, underscoring the limitations of adapting just the encoder and mapping layers without additional trainable parameters in LLM. Fine-tuning the encoder alongside LoRA yields the best results, as it jointly optimizes foundational and task-specific representations. In contrast, the Mapper Only configuration, which fine-tunes only the mapping layer, performs poorly due to the fixed weights of the audio encoder, which hinder the model’s ability to adapt to heart and lung sounds. Performance deteriorates further in the Frozen Encoder setting, reinforcing the critical role of encoder fine-tuning in effective learning.

VeRA also lags behind LoRA, as its vector-based residual adapters offer less flexibility and capacity for adaptation. Although VeRA introduces task-specific modifications, its updates fail to capture complex relationships as effectively as LoRA's low-rank matrix-based updates, which better fine-tune key layers of the model for the task at hand.

\textbf{Comparison of PEFT Methods.} 

Table~\ref{peft-methods} compares the performance of various PEFT methods. LoRA consistently outperforms prefix tuning, prompt tuning, and P-tuning. While prefix tuning and prompt tuning focus on modifying lightweight input-related parameters, their limited scope constrains their ability to capture complex relationships in the data. P-tuning offers slightly greater flexibility but still falls short of LoRA. By incorporating trainable parameters into the deeper layers of the model, LoRA enables more effective adaptation to task-specific requirements, resulting in superior performance.

\begin{table}[t]
\centering
\caption{Performance of Different Mapping Types.}
\label{mapper}
\scriptsize
\begin{tabular}{lcc}
\toprule
\textbf{Mapping Type} & \textbf{BertScore} & \textbf{METEOR} \\
\midrule
Linear       & 85.2 & 76.9 \\
MLP          & 85.3 & 76.3 \\
Transformer  & \textbf{86.2} & \textbf{77.5} \\
\bottomrule
\end{tabular}
\end{table}

\textbf{Effect of LoRA Rank.} 
Effect of LoRA Rank. Table~\ref{lora-rank} analyzes the impact of varying the LoRA rank \textit{r}. A rank of 8 achieves an optimal balance between efficiency and performance. Lower ranks fail to provide adequate capacity to capture complex relationships in the data, while higher ranks introduce excessive complexity, leading to a slight decline in performance.

\textbf{Mapping Types and Audio Encoder Types.}
Tables~\ref{mapper} and~\ref{encoder-comparison-table} evaluate the impact of mapping types and audio encoder types on model performance. Among the mapping types, the Transformer-based approach achieves the highest BertScore and METEOR values, demonstrating its ability to model complex dependencies and capture long-range interactions through self-attention. This capability makes it particularly effective for sequential data like audio. In contrast, linear mappings are restricted to basic transformations, while MLP mappings, despite their non-linear nature, lack the expressiveness needed to model intricate feature relationships effectively.

\begin{table}[t]
\centering
\caption{Performance of Audio Encoder Types.}
\label{encoder-comparison-table}
\scriptsize
\begin{tabular}{lcc}
\toprule
\textbf{Encoder Type} & \textbf{BertScore} & \textbf{METEOR} \\
\midrule
CLAP      & 76.8    & 72.4      \\
OPERA-GT  & 79.5 & \textbf{81.9} \\
OPERA-CE  & \textbf{86.2} & 77.5 \\
\bottomrule
\end{tabular}
\end{table}

\looseness=-1
For encoder types, OPERA-CE achieves the best overall performance, with the highest BertScore and competitive METEOR values. This highlights the effectiveness of contrastive pre-training in aligning audio and text representations, a critical requirement for tasks like audio-QA. OPERA-GT also performs well, particularly in METEOR, likely due to generative pre-training’s focus on producing coherent outputs. However, its slightly lower BertScore indicates less precise cross-modal alignment. The CLAP encoder, pre-trained for broader language-audio tasks, lags behind in both metrics, reflecting its general-purpose design and reduced specialization for audio-health QA tasks.

\section{Conclusion}
\looseness=-1
This work introduces CaReAQA, an audio-language model specifically designed  for diagnostic reasoning in cardiac and respiratory auscultation. By integrating a foundational audio model with the advanced reasoning capabilities of a large language model, CaReAQA significantly outperforms general-purpose baselines on diverse datasets, demonstrating its proficiency in generating accurate and clinically relevant diagnostic answers. Specifically, a crucial contribution is the development of the CaReSound benchmark, a diverse dataset designed to advance open-ended question-answering in health diagnostics. Our experimental results show the model’s strong generalization across both open-ended and closed-ended tasks in CaReSound, highlighting its potential for real-world clinical applications. Future work will expand the dataset, improve multimodal fusion, enhance interpretability, and explore regulatory pathways for real-world deployment.

\section*{Acknowledgments}
This work was supported by the NGF AiNed Fellowship Grant of A.S. We also acknowledge the use of the Dutch National Supercomputer Snellius for essential computational tasks.

\bibliography{main}
\clearpage

\onecolumn
\appendix

\section{Samples from CaReSound Dataset}
\label{apd:training_data}
This section provides examples from the CaReSound dataset. The dataset is designed to facilitate research in medical diagnostics by providing paired audio recordings and text-based question-answer data. Table~\ref{qa-sample-table} includes representative examples that demonstrate the dataset's structure and content. These examples capture the dataset's range, encompassing various types of questions, answers, and corresponding audio samples, which are essential for training and evaluating models in clinical settings.

\begin{longtable}{p{5cm} p{6cm} m{3cm}}
\caption{Randomly Selected Examples of Audio Question-Answer Pairs.}
\label{qa-sample-table} \\
\toprule
\textbf{Question} 
  & \textbf{Answer} 
  & \textbf{Audio} \\
\midrule
\endfirsthead
\toprule
\textbf{Question} 
  & \textbf{Answer} 
  & \textbf{Audio} \\
\midrule
\endhead
Q: Is the murmur localized to just one valve area? 
  & A: No, the murmur is heard at the aortic, mitral, pulmonic, and tricuspid valve locations.
  & \includegraphics[width=3cm]{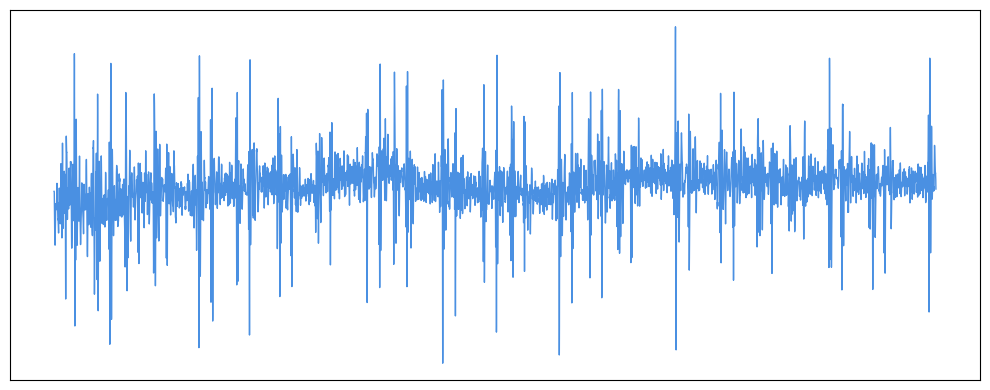} \\[10pt]
Q: What type of sounds were identified during auscultation?
  & A: Inspiratory Crackles and Expiratory Wheezes.
  & \includegraphics[width=3cm]{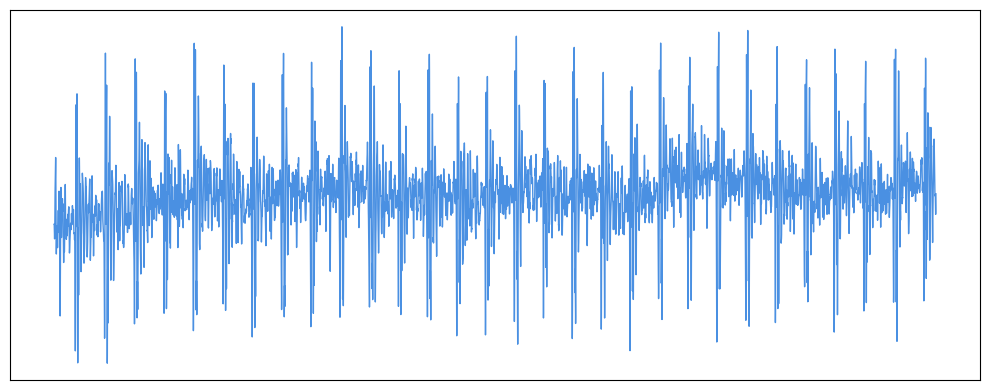} \\[10pt]
Q: Were any abnormal lung sounds detected over the trachea in the diagnosis of COPD?
  & A: No abnormal lung sounds like crackles or wheezes were detected over the trachea in the diagnosis of COPD.
  & \includegraphics[width=3cm]{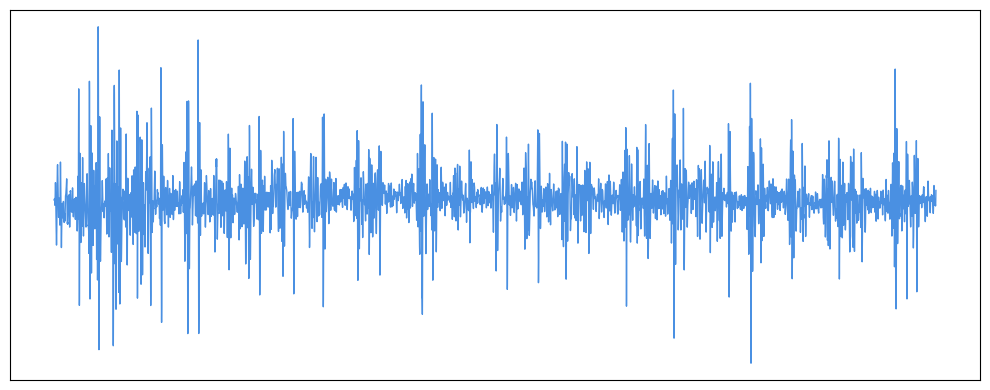} \\[10pt]
Q: Is the presence of fine crackles indicative of any specific lung condition?
  & A: Yes, fine crackles can be indicative of conditions such as pulmonary fibrosis or heart failure.
  & \includegraphics[width=3cm]{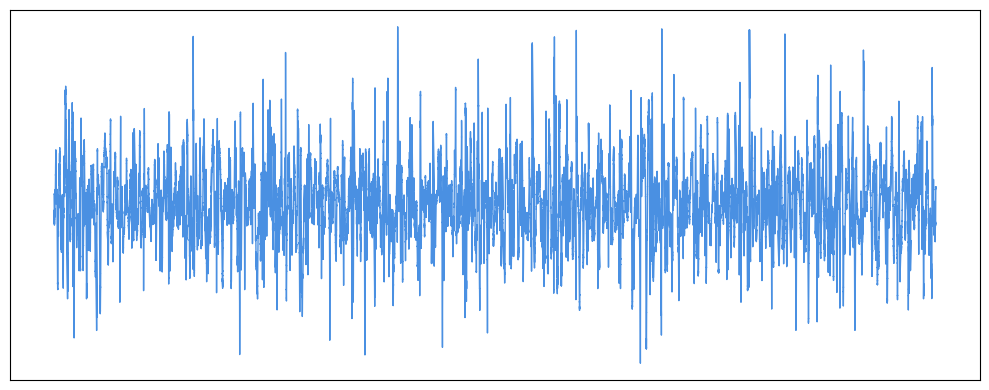} \\[10pt]
  Q: What types of sounds were auscultated? & A: Inspiratory Crackles and Expiratory Wheezes 
  & \includegraphics[width=3cm]{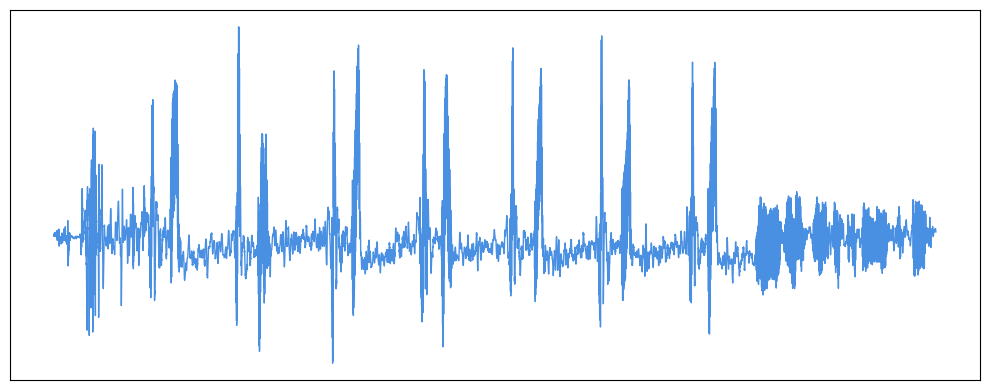} \\[10pt]
  Q: In which part of the chest were crackles detected? & A: Crackles were detected in the lateral right part of the chest. & \includegraphics[width=3cm]{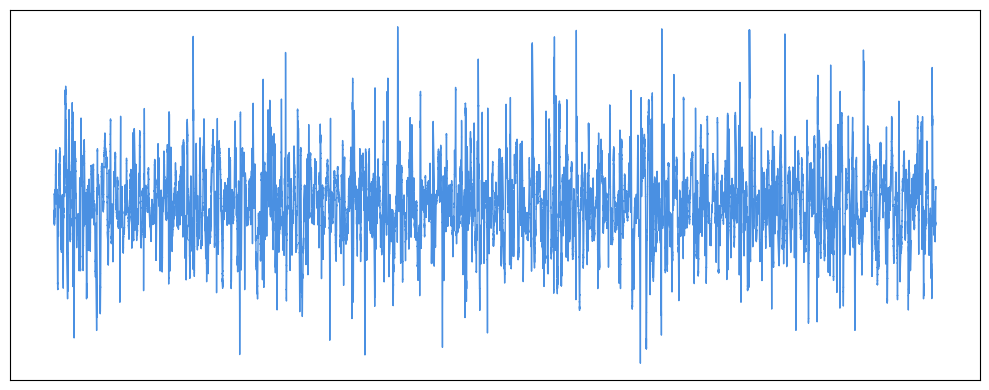} \\[10pt]
  Q: Are there any signs of respiratory distress in the auscultation results? & A: No, the results are consistent with normal respiratory function. &
  \includegraphics[width=3cm]{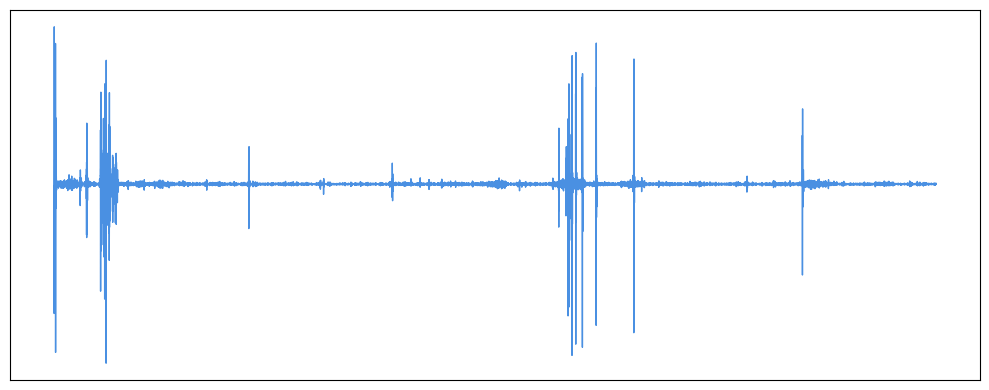} \\[10pt]
  Q: Are there any abnormal heart sounds present in the cardiac auscultation findings? & A:	No, there are no abnormal heart sounds present in the cardiac auscultation findings. &
  \includegraphics[width=3cm]{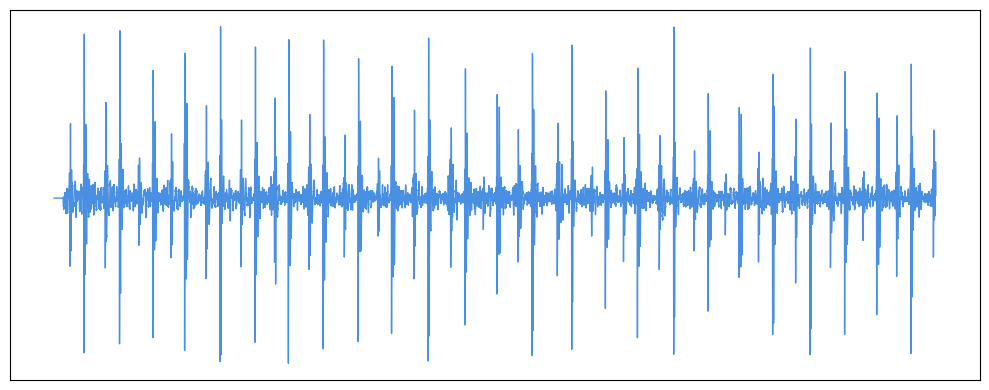} \\[10pt] 
\bottomrule
\end{longtable}


\section{Prompt for Accuracy Evaluation with GPT4o\citep{openai2024gpt4technicalreport}}
\label{apd:accuracy-eval}

To ensure consistency and reliability in evaluating open-ended question-answering tasks, a dedicated prompt was used for assessing the similarity between the ground-truth answers and the model predictions. This evaluation framework aims to standardize the process by enforcing a binary output—either ``Yes'' or ``No''—indicating whether the prediction matches the ground truth. Such an approach is critical for minimizing ambiguity in model assessment and ensuring reproducibility in accuracy evaluations.

\begin{tcolorbox}[
    colframe=black,
    title=\footnotesize{Prompt for Accuracy Evaluation}
]
Your job is to evaluate if the ground-truth and prediction are same/similar. \\ \\
Provide only Yes or No answer as JSON of the following structure:
\{\textquotesingle answer\textquotesingle: \textquotesingle \textquotesingle\} without any explanation. \\ 

Ground-truth: \texttt{\$\{ground\_truth\}}\\
Prediction: \texttt{\$\{prediction\}}
\label{box:accuracy-eval}
\end{tcolorbox}

\section{Prompt Example for QA Generation}
\label{apd:appendix-prompt}

Generating clinically relevant QA pairs is integral to the study, as it ensures the alignment of model outputs with real-world diagnostic scenarios. The prompt provided here instructs the model to simulate the reasoning process of a clinician interpreting respiratory auscultation findings. By constraining the generated QA pairs to metadata-derived information, the prompt eliminates potential bias or reliance on external assumptions, guaranteeing that the outputs remain clinically applicable and scientifically sound.

\begin{tcolorbox}[
    colframe=black,
    title=\footnotesize{Prompt for Generating Clinically Relevant QA Pairs}
]
You are a clinician tasked with interpreting respiratory auscultation findings. \\ \\
Based on the given conditions, your job is to generate at least 3 question-answer (QA) pairs that are clinically relevant. Note that the questions and answers should be based only on the provided metadata and should not include any external assumptions. \\ \\
Your output should follow this structure: \\ \\

\{
  \texttt{"QAs"}: [
    \{\texttt{"question"}: "...", \texttt{"answer"}: "..."\},
    \{\texttt{"question"}: "...", \texttt{"answer"}: "..."\},
    \{\texttt{"question"}: "...", \texttt{"answer"}: "..."\}
  ]
\}
\label{box:clinician-qa}
\end{tcolorbox}

\clearpage

\section{Language Model Performance Comparison}
\label{apd:model_performance}
This section compares the performance of various large language models (LLMs) on the CaReSound dataset. The evaluation focuses on metrics such as BERTScore and METEOR, which are widely used to measure text similarity and relevance. The results of these evaluations, presented in Table \ref{model_comparison}. 
The results demonstrate that LLaMA3.2-3B outperforms competing models, demonstrating its capability to handle the intricacies of audio-QA tasks. These findings underline the importance of model selection in achieving high accuracy and reliability in domain-specific applications.

\begin{table}[htbp]
\centering
\small
\caption{Comparison of different LLMs across datasets.}
\label{model_comparison}
\begin{tabular}{lcc}
    \toprule
    \textbf{Method} & \textbf{BertScore} & \textbf{METEOR} \\
    \midrule
    Gemma2-2b  \citep{gemma2024}    & 73.8 & 64.6 \\
    Qwen1.5    \citep{chu2023qwen}    & 77.3 & 64.9 \\
    SmolLM-1.7B  \citep{allal2024SmolLM2}  & 81.3 & 66.7 \\
    DeepSeek-R1-Distill-1.5B \citep{guo2025deepseek} & 80.0 &  72.0 \\
    LLaMA3.2-3B \citep{llama2023}  & \textbf{86.2} & \textbf{77.5} \\
    \bottomrule
\end{tabular}
\end{table}

\section{Comparison of Instruction-Tuned Language Models}
\label{apd:instruction-tuned}

We compared multiple large language models—namely gemma-2b-it \citep{gemma2024}, Qwen2-1.5B-Instruct\citep{chu2023qwen}, and SmolLM2-1.7B-Instruct\citep{allal2024SmolLM2}, Llama-3.2-3B-Instruct\citep{llama2023}—to the base Llama-3.2-3B under identical fine-tuning conditions tailored to audio-QA. As shown in Table~\ref{instruct-models-table}, while gemma-2b-it demonstrated occasional benefits, the base model consistently matched or exceeded the performance of all instruction-tuned options once domain-specific training was applied. These results indicate that directly refining the core model can be more effective than relying on broad instruction-tuning strategies for specialized tasks. Consequently, we adopted the base Llama-3.2-3B for CaReAQA.

\begin{table*}[ht!]
\centering
\caption{Comparison of Instruct Models on BERTScore (BertS) and METEOR across considered datasets (ICBHI, CIRCOR, KAUH, SPR, and ZCH).}
\label{instruct-models-table}
\resizebox{\textwidth}{!}{%
\begin{tabular}{l
                cc
                cc
                cc
                cc
                cc
                cc}
    \toprule
    & \multicolumn{2}{c}{\textbf{ICBHI}} 
    & \multicolumn{2}{c}{\textbf{CIRCOR}} 
    & \multicolumn{2}{c}{\textbf{KAUH}} 
    & \multicolumn{2}{c}{\textbf{SPR}} 
    & \multicolumn{2}{c}{\textbf{ZCH}}  \\
    \cmidrule(lr){2-3}
    \cmidrule(lr){4-5}
    \cmidrule(lr){6-7}
    \cmidrule(lr){8-9}
    \cmidrule(lr){10-11}
    \cmidrule(lr){12-13}
    \textbf{Model}
    & \textbf{BertS} & \textbf{METEOR}
    & \textbf{BertS} & \textbf{METEOR}
    & \textbf{BertS} & \textbf{METEOR}
    & \textbf{BertS} & \textbf{METEOR}
    & \textbf{BertS} & \textbf{METEOR} \\
    \midrule
    Llama-3.2-3B-Instruct \citep{llama2023}
        & 83.1 & 75.0
        & 86.0 & 80.2
        & 68.0 & 54.2
        & 79.2 & 72.5
        & 81.3 & 75.6\\
    Qwen2-1.5B-Instruct \citep{chu2023qwen}
        & 80.2 & 74.5
        & 78.6 & 70.9
        & 55.6 & 34.5
        & 75.5 & 70.4
        & 76.8 & 71.5\\
    SmolLM2-1.7B-Instruct \citep{allal2024SmolLM2}
        & 81.0 & 73.2
        & 78.1 & 67.3
        & 58.0 & 46.4
        & 72.9 & 68.1
        & 74.8 & 70.4 \\
    Gemma-2b-it \citep{gemma2024}
        & 83.9 & 76.2
        & 88.5 & 81.8
        & 68.4 & 59.3
        & 80.8 & 74.2
        & 82.9 & 77.2 \\
    \bottomrule
\end{tabular}%
}
\end{table*}

\section{Failure Case Analysis}
\label{apd:failure_cases}

\noindent Understanding the limitations of our model is crucial for improving its clinical applicability. While CaReAQA demonstrates strong performance across diverse datasets, certain cases remain challenging. These failure cases primarily fall into three categories: (i) conditions with overlapping auscultation patterns, making differentiation difficult, (ii) misclassifications of less common diseases due to limited training samples, and (iii) ambiguous or noisy recordings where diagnostic certainty is inherently low.

Table~\ref{tab:failure_cases} presents representative failure cases where the model's predictions deviated from the ground truth, highlighting areas for further improvement. These cases illustrate the challenges of auscultation-based diagnostics, especially when conditions share similar acoustic features.

\begin{table}[htbp]
\centering
\small
\caption{Representative Failure Cases in Diagnostic Predictions}
\label{tab:failure_cases}
\begin{tabular}{p{5.5cm} p{4.5cm} p{4.5cm}}
    \toprule
    \textbf{Question} & \textbf{Ground Truth} & \textbf{Model Prediction} \\
    \midrule
    At which chest location were crackles heard? & Crackles were heard at the lateral left chest location. & Crackles were heard on the anterior right chest location. \\
    Which condition is diagnosed based on the detected murmur? & Aortic stenosis. & Mitral regurgitation. \\
    Does the patient have an underlying cardiac condition? & Yes, the findings suggest heart failure. & No significant cardiac abnormality detected. \\
    Is there evidence of a cardiac anomaly in the auscultation? & No, there is no evidence of a cardiac anomaly; the findings are normal. & Yes, there is evidence of a cardiac anomaly. \\
    What type of systolic murmur is present? & A holosystolic, low-pitched, blowing murmur with a plateau shape. & A holosystolic, high-pitched, harsh murmur. \\
    What is the grade of the murmur?&The murmur is graded I/VI.&The murmur is graded III/VI.\\
    \bottomrule
\end{tabular}
\end{table}

\end{document}